\documentclass[runningheads]{llncs}
\usepackage{graphicx}

\usepackage{tikz}
\usepackage{comment} 
\usepackage{amsmath,amssymb} 
\usepackage{color}

\usepackage{threeparttable}
\usepackage{hyperref}

\newcommand{\etal}{{et al}.\@ }

\usepackage{environ}
\newcommand{\acksection}{\section*{Acknowledgments}}
\NewEnviron{acks}{%
  \acksection
  \BODY
}

\begin{document}
\pagestyle{headings}
\mainmatter
\def\ECCVSubNumber{916}  

\title{Occlusion-Aware Depth Estimation with Adaptive Normal Constraints} 

\titlerunning{Occlusion-Aware Depth Estimation with Adaptive Normal Constraints}
%
\author{Xiaoxiao Long\inst{1} \and
Lingjie Liu\inst{2} \and
Christian Theobalt\inst{2} \and Wenping Wang\inst{1}}
\authorrunning{X. Long et al.}
%
\institute{The University of Hong Kong \\ \email{\{xxlong,wenping\}@cs.hku.hk} \and
Max Planck Institute for Informatics \\
 \email{\{lliu,theobalt\}@mpi-inf.mpg.de} }
\maketitle

\begin{abstract}
We present a new learning-based method for multi-frame depth estimation from a color video, which is a fundamental problem in scene understanding, robot navigation or handheld 3D reconstruction. While recent learning-based methods estimate depth at high accuracy, 3D point clouds exported from their depth maps often fail to preserve important geometric feature (e.g., corners, edges, planes) of man-made scenes. Widely-used pixel-wise depth errors do not specifically penalize inconsistency on these features. 
These inaccuracies are particularly severe when subsequent depth reconstructions are accumulated in an attempt to scan a full environment with man-made objects with this kind of features. 
Our depth estimation algorithm therefore introduces a {\em Combined Normal Map (CNM)} constraint, which is designed to better preserve high-curvature features and global planar regions. 
In order to further improve the depth estimation accuracy, we introduce a new occlusion-aware strategy that aggregates initial depth predictions from multiple adjacent views into one final depth map and one occlusion probability map for the current reference view.
Our method outperforms the state-of-the-art in terms of depth estimation accuracy, and preserves essential geometric features of man-made indoor scenes much better than other algorithms. Code is available: \url{https://github.com/xxlong0/CNMNet}.

\end{abstract}

\section{Introduction}
Dense multi-view stereo is one of the fundamental problems in computer vision with decades of research, e.g.~\cite{Seitz:2006:CEM:1153170.1153518,hosni2012fast,bleyer2011patchmatch,yang2012non,hosni2009local,yoon2006adaptive}. Algorithms vastly differ in their assumptions on input (pairs of images, multi-view images etc.) or employed scene representations (depth maps, point clouds, meshes, patches, volumetric scalar fields etc.)~\cite{Furukawa:2015:MST:2864699.2864700}.

In this paper, we focus on the specific problem of multi-view depth estimation from a color video from a single moving RGB camera~\cite{Furukawa:2015:MST:2864699.2864700}. Algorithms to solve this problem have many important applications in computer vision, graphics and robotics. They empower robots to avoid collisions and plan paths using only one onboard color camera~\cite{alvarez2016collision}. Such algorithms thus enable on-board depth estimation from mobile phones to properly combine virtual and real scenes in augmented reality in an occlusion-aware way. 
Purely color-based solutions bear many advantages over RGB-D based approaches~\cite{Zollhoefer2018RecoSTAR} since color cameras are ubiquitous, cheap, consume little energy, and work nearly in all scene conditions.

Depth estimation from a video sequence is a challenging problem. Traditional methods~\cite{hosni2012fast,bleyer2011patchmatch,yang2012non,hosni2009local,yoon2006adaptive} achieve impressive results but struggle on important scene aspects: texture-less regions, thin structures, shape edges and features, and non-Lambertian surfaces.
Recently there are some attempts to employ learning techniques to this problem~\cite{Kendall_2017_ICCV,huang2018deepmvs,wang2018mvdepthnet,yao2018mvsnet,im2019dpsnet}. These methods train an end-to-end network typically with a pixel-wise depth loss function. They lead to significant accuracy improvement in depth estimation compared to non-learning-based approaches. However, most of these works fail to preserve prominent features of 3D shapes, such as corners, sharp edges and planes, because they use only depth for supervision and their loss functions are therefore not built to preserve these structures. 
This problem is particularly detrimental when reconstructing indoor scenes with man-made objects or regular shapes, as shown in Fig.~\ref{fig:depth_compare_single}.
Another problem is the performance degradation caused by the depth ambiguity in the occluded region, which has been ignored by most of the existing works. 

We present a new method for depth estimation with a single moving color camera that is designed to preserve important local features (edges, corners, high curvature features) and planar regions. It takes one video frame as {\em reference image} and uses some other frames as {\em source images} to estimate depth in the reference frame. Pairing the reference image with each source image, we first build an initial 3D cost volume from each pair via plane-sweeping warping. Subsequent cost aggregation for each initial cost volume yields an initial depth maps for each image pair (e.g. the reference image and a source image). 
Then we employ a new occlusion-aware strategy to combine these initial depth maps into one final reference-view depth map, along with an occlusion probability map. 

Our first main contribution is a new structure preserving constraint enforced during training. 
It is inspired by learning-based monocular depth estimation methods using normal constraints for structure preservation.
GeoNet~\cite{qi2018geonet} enforces a surface normal constraint, but their results have artifacts due to noise in the ground truth depth.
Yin \etal~\cite{Yin2019enforcing} propose a global geometric constraint, called \emph{virtual normal}. However, they cannot preserve intrinsic geometric features of real surfaces and local high-curvature features. 
We therefore propose a new \emph{Combined Normal Map (CNM)} constraint, attached to local features for both local high-curvature regions and global planar regions. For training our network, we use a differentiable least squares module to compute normals directly from the estimated depth and use the \emph{CNM} as ground truth in addition to the standard depth loss. Experiments in Section~\ref{surface_normal_compare} show that the use of this novel \emph{CNM} constraint significantly improves the depth estimation accuracy and outperforms those approaches that use only local or global constraints.

Our second contribution is a new neural network that combines depth maps predicted with individual source images into one final reference-view depth map, together with an occlusion probability map. It uses a novel occlusion-aware loss function which assigns higher weights to the non-occluded regions. Importantly, this network is trained without any occlusion ground truth.


We experimentally show that our method significantly outperforms the state-of-the-art multi-view stereo from monocular video, both quantitatively and qualitatively. Furthermore, we show that our depth estimation algorithm, when integrated into a fusion-based handheld 3D scene scanning approach, enables interactive scanning of man-made scenes and objects in much higher shape quality (See Fig.~\ref{fig:video_recon}).

\section{Related Work}

{\bf Multi-view stereo.} MVS algorithms~\cite{hosni2012fast,bleyer2011patchmatch,yang2012non,hosni2009local,yoon2006adaptive} are able to reconstruct 3D models from images under the assumptions of known materials, viewpoints, and lighting conditions. The MVS methods vary significantly according to different scene representations, but the typical workflow is to use hand-crafted photo-consistency metrics to build a cost volume and do the cost aggregation and estimate the 3D geometry from the aggregated cost volume. These methods tend to fail for thin objects, non-diffuse surfaces, or the objects with insufficient features.

\noindent
{\bf Learning-based depth estimation.} 
Recently some learning-based methods achieve compelling results in depth estimation. They can be categorized into four groups: i) single-view depth estimation~\cite{eigen2014depth,liu2015deep,li2018megadepth,fu2018deep}; ii) two-view stereo depth estimation~\cite{zbontar2015computing,luo2016efficient,ummenhofer2017demon,chang2018pyramid,Kendall_2017_ICCV}; iii) multi-view stereo depth estimation~\cite{huang2018deepmvs,yao2018mvsnet,im2019dpsnet}; iv) depth estimation from a video sequence~\cite{wang2018mvdepthnet,liu2019neural}. Single-view depth estimation is an ill-posed problem due to inherent depth ambiguity. On the other hand, two-view and multi-view depth estimation~\cite{Furukawa:2015:MST:2864699.2864700} is also challenging due to the difficulties in dense correspondence matching and depth prediction in featureless or specular regions. 
Some two-view~\cite{zbontar2015computing,luo2016efficient,chang2018pyramid,Kendall_2017_ICCV} and multi-view~\cite{huang2018deepmvs,wang2018mvdepthnet,yao2018mvsnet,im2019dpsnet} depth estimation algorithms have produced promising results in accuracy and speed by integrating cost volume generation, cost volume aggregation, disparity optimization and disparity refinement in an end-to-end network. 

Depth estimation from video frames is becoming popular with the high demand in emerging areas, such as AR/VR, robot navigation, autonomous driving and view-dependent realistic rendering.  Wang \etal~\cite{wang2018mvdepthnet} design a real-time multi-view depth estimation network. Liu \etal~\cite{liu2019neural} propose a Bayesian filtering framework to use frame coherence to improve depth estimation. Since these methods enforce depth constraint only, they often fail to preserve important geometric feature (e.g., corners, edges, planes) of man-made scenes. 

\noindent
{\bf Surface normal constraint.} Depth-normal consistency has been explored before for the depth estimation task~\cite{eigen2015predicting,zhang2018deepdepth,qi2018geonet,Yin2019enforcing}. 
Eigen \etal~\cite{eigen2015predicting} propose a neural network with three decoders to separately predict depth, surface normal and segmentation. Zhang \etal~\cite{zhang2018deepdepth} enforce the surface normal constraint for depth completion. Qi \etal~\cite{qi2018geonet} incorporate geometric relations between depth and surface normal by introducing the depth-to-normal and normal-to-depth networks. The performance of these methods suffers from the noise in surface normal stemming from ground truth depth. 
Yin \etal~\cite{Yin2019enforcing} propose a global geometric constraint, called \emph{virtual normal}, which is defined as the normal of a virtual surface plane formed by three randomly sampled non-collinear points with a large distance. The \emph{virtual normal} is unable to faithfully capture the geometric features of real surfaces and the local high-curvature features. Kusupati \etal~\cite{kusupati2019normal} use the sobel operator to calculate the spatial gradient of depth and enforce the consistency of the spatial gradient and the normal in the pixel coordinate space. However the gradient calculated by the sobel operator is very sensitive to local noise, and causes obvious artifacts in the surface normal calculated from estimated depth (See Fig.~\ref{fig:normal_compare_geonet} for the visual result).

\noindent
{\bf Occlusion handling for depth estimation.} Non-learning methods~\cite{kang2001handling,egnal2002detecting,xu2008stereo,xiao2005motion,min2008cost,humayun2011learning} use post-processing, such as left-right consistency check~\cite{egnal2002detecting}, to handle the occlusion issue. There are some learning-based methods for handling occlusion as well. 
Ilg \etal~\cite{ilg2018occlusions} and Wang \etal~\cite{wang2019local} predict occlusions by training a neural network on a synthetic dataset in a supervised manner. Qiu \etal~\cite{qiu2019deeplidar} directly learn an intermediate occlusion mask for a single image; it has no epipolar geometry guarantee and just drives network to learn experienced patterns. We propose an occlusion-aware strategy to jointly refine depth prediction and estimate an occlusion probability map for multi-view depth estimation without any ground truth occlusion labels. 

\section{Method}
\begin{figure*}[t]
    \centering
    \includegraphics[width=\textwidth]{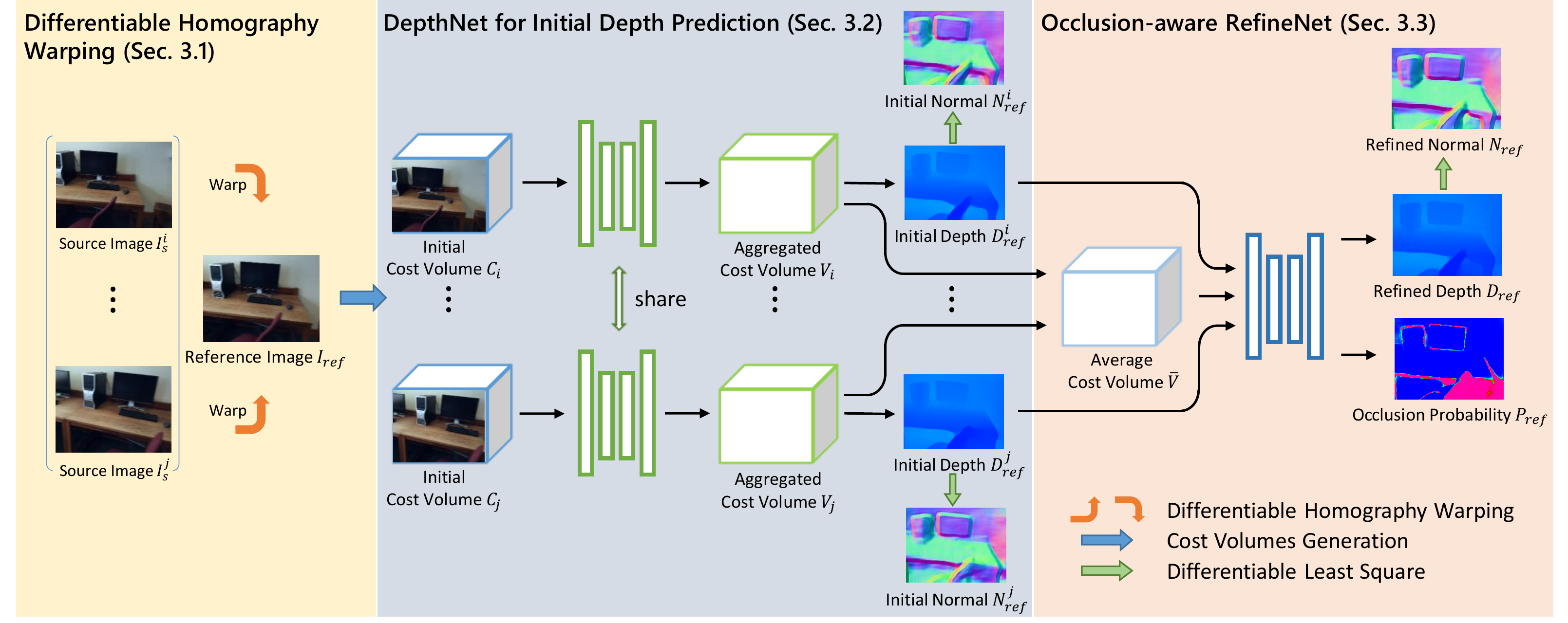}
    \caption{Overview of our method. The network input consists of one reference image and $n (n=2,4,\dots)$ source images. With homography warping, each source image is combined with the reference image as a pair to generate a cost volume. Then the cost volume is fed into the \emph{DepthNet} to generate an initial depth map, with $\ell_1$ depth supervision and the constraint of \emph{Combined Normal Map}. Finally, with the $n$ initial depth maps and the average cost volume of the aggregated cost volumes as input, the \emph{RefineNet} employs an occlusion-aware loss to produce the final reference-view depth map and an occlusion probability map, with again the supervision by the depth and the \emph{CNM} constraints.
    }
    \label{fig:pipeline}
\end{figure*}

In this section we describe the proposed network, as outlined in Fig.~\ref{fig:pipeline}. Our pipeline takes the frames in a local time window in the video as input. Note that in our setting the video frame rate is 30 fps and we sample video frames with the interval of 10. 

For the sake of simplicity in exposition, we suppose that the time window size is 3. 
We take the middle frame as the reference image ${I}_{ref}$, and the two adjacent images as the two source images $\{I_s^1, I_s^2 \}$ which are supposed to have sufficient overlap with the reference image. 
Our goal is to compute an accurate depth map from the view of the reference image.

We first outline our method. With a differentiable homography warping operation~\cite{yao2018mvsnet}, all the source images are first warped into a stack of different fronto-parallel planes of the reference camera to form an initial 3D cost volume (see Section~\ref{sweepingplane}). Next, cost aggregation is applied to the initial cost volume to rectify any incorrect cost values and then an initial depth map is extracted from the aggregated cost volume. Besides pixel-wise depth supervision over the initial depth map, we also enforce a novel local and global geometric constraint, namely \emph{Combined Normal Map (CNM)}, for training the network to produce superior results (see Section~\ref{depthnet}). We then further improve the accuracy of depth estimation by applying a new occlusion-aware strategy to aggregate the depth predictions from different adjacent views into one depth map for the reference view, with an occlusion probability map (see Section~\ref{refinenet}). The details of each step will be elaborated in the subsequent sections.

\subsection{Differentiable Homography Warping}
\label{sweepingplane}
Consider the reference image ${I}_{ref}$ and one of its source images, denoted ${I}_{s}^{i}$. We warp ${I}_{s}^{i}$ into fronto-parallel virtual planes of ${I}_{ref}$ to form an initial cost volume. 
Similar to~\cite{huang2018deepmvs,yao2018mvsnet}, we first uniformly sample $D$ depth planes at depth $d_n$ in a range $[d_{min}, d_{max}]$, $n=1, 2, \dots, D$. 
The mapping from the the warped ${I}_{s}^{i}$ to the $n^{th}$ virtual plane of ${I}_{ref}$ at depth $d_n$ is determined by a planar homography transformation $H_{n}$, following the classical plane-sweeping stereo \cite{gallup2007real}. 
\begin{equation}
u'_{n}\sim H_{n}u,u'_{n}\sim \mathbf{K}[{\mathbf{R}}_{s,i}|{\mathbf{t}}_{s,i}]\begin{bmatrix}
({\mathbf{K}}^{-1}u)d_{n} \\
1 
\end{bmatrix},
\end{equation}
where $u$ is the homogeneous coordinate of a pixel in reference image, $u'_{n}$ is projected coordinate of $u$ in source image ${I}_{s}^{i}$ with virtual plane $d_{n}$. $K$ denotes the intrinsic parameters of the camera, $\{ \mathbf{R}_{s,i}, \mathbf{t}_{s,i} \}$ are the rotation and the translation of the source image ${I}_{s}^{i}$ relative to the reference image $I_{ref}$.

Next, we measure the visual consistency of the warped ${I}_{s}^{i}$ and ${I}_{ref}$ at depth $d_n$ and build a cost volume ${C}_i$ with the size of $W \times H \times D$, where $W, H, D$ are the image width, image height and the number of depth planes, respectively. Unlike previous works~\cite{yao2018mvsnet,liu2019neural} that use extracted feature maps of an image pair for warping and building a 4D cost volume, here we use the image pair directly to avoid the memory-heavy and time-consuming 3D convolution operation on a 4D cost volume.

\subsection{DepthNet for Initial Depth Prediction}
\label{depthnet}
Similar to recent learning-based stereo~\cite{Kendall_2017_ICCV} and MVS~\cite{yao2018mvsnet,im2019dpsnet} methods, after getting cost volumes $\{ {C}_i \}_{i=1,2}$ from image pairs $\{ I_{ref}, I_{s}^{i} \}_{i=1,2}$, we first use the neural network \emph{DepthNet} to do cost aggregation for each ${C}_i$, which rectifies the incorrect values by aggregating the neighbouring pixel values. Note that we feed $C_i$ stacked with $I_s$ together into the \emph{DepthNet} in order to make use of more detailed context information as~\cite{wang2018mvdepthnet} does. We denote the aggregated cost volume as $V_i$. 
Next, we retrieve the initial depth map $D_{ref}^i$ from the aggregated cost volume $V_i$ using a 2D convolution layer. 
Note that two initial depth maps, $D_{ref}^1$ and $D_{ref}^2$, are generated for the reference view $I_{ref}$.

We train the network with depth supervision. With only depth supervision, the point cloud converted from the estimated depth does not preserve regular features, such as sharp edges and planar regions. We therefore propose to also enforce the normal constraint for further improvement. Note that, instead of just using  local surface normal~\cite{qi2018geonet} or just a global virtual normal~\cite{Yin2019enforcing}, we use the so called \emph{Combine Normal Map} (CNM) that combines the local surface normal and the global planar structural feature in an adaptive manner.

{\bf Pixel-wise Depth Loss}
We use a standard pixel-wise depth map loss as follows,
\begin{equation}
\label{depthloss}
{l}_{id}=\frac{1}{\left| \mathcal{Q} \right|} \sum_{q \in \boldsymbol{Q}} \left\|\hat{D}(q)-D_{i}(q)\right\|_{1}
\end{equation}
where $\boldsymbol{Q}$ is the set of all pixels that are valid in ground truth depth, $\hat{D}(q)$ is the ground truth depth value of pixel $q$, and ${D}_{i}(q)$ is the initial estimated depth value of pixel $q$.

\begin{figure*}[h]
    \centering
    \includegraphics[width=0.9\linewidth]{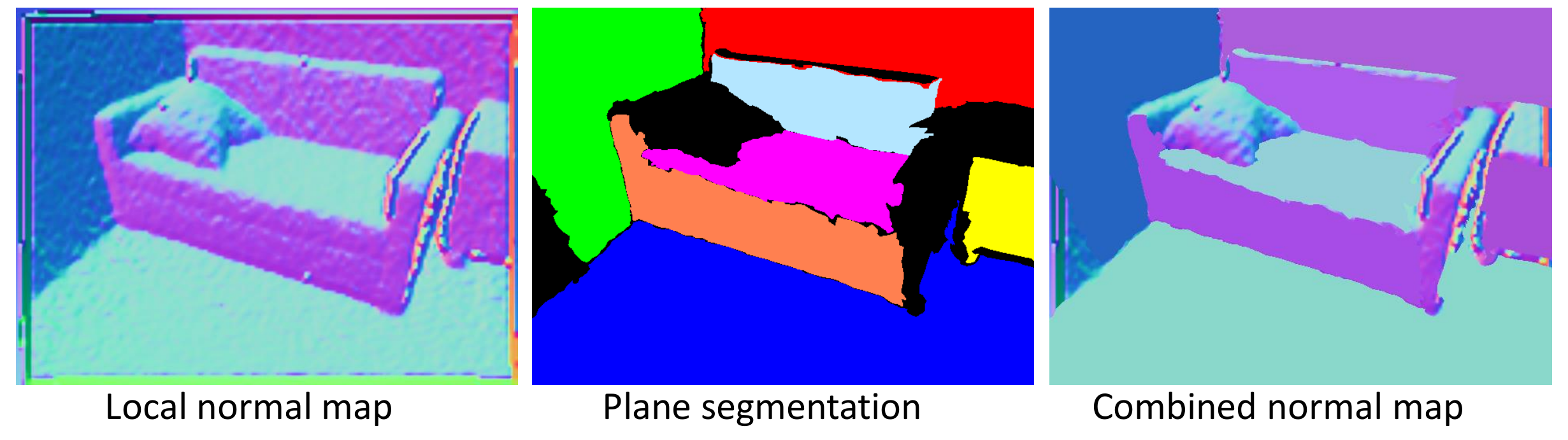}
    \caption{Visual comparison of local normal map and combined normal map.}
    \label{fig:CNM_case}
\end{figure*}

{\bf Combined Normal Map}
In order to preserve both local and global structures of scenes, we introduce the \emph{Combined Normal Map (CNM)} as ground truth for the supervision with the normal constraint. To obtain this normal map, we first use PlaneCNN~\cite{Liu_2019_CVPR} to extract planar regions, such as walls, tables, and floors. Then we apply the local surface normals to non-planar regions, and use the mean of the surface normals in a planar region as the assigned to the region. The visual comparison of the local normal map and the \emph{CNM} can be seen in Fig.~\ref{fig:CNM_case}.

The key insight here is to use local surface normal to capture rich local geometric features in the high-curvature regions and to employ the average normal to filter out noise in the surface normals for the planar regions to preserve global structures. In this way, the \emph{CNM} significantly improves the depth prediction and the recovery of good 3D structures of the scene, compared to using only local or global normal supervision (see the ablation study in Section~\ref{Ablation}).

{\bf Combined Normal Loss} We define the loss on the CNM as follows: 
\begin{equation}
\label{normalloss}
{l}_{in} = -\frac{1}{\left| \mathcal{Q} \right|} \sum_{q \in \boldsymbol{Q}}
\hat{N}(q) \cdot {N}_{i}(q),
\end{equation}
where $\boldsymbol{Q}$ is the set of valid ground truth pixels, $\hat{N}(q)$ is the combined normal of pixel $q$, and ${N}_{i}(q)$ is the surface tangent normal of the 3D point corresponding to the pixel $q$, both normalized to unit vectors. 

To obtain an accurate depth map and preserve geometric features, we combine the Pixel-wise Depth Loss and the Combined Normal Loss together to supervise the network output. The overall loss is:
\begin{equation}
\label{depthnetloss}
{l}_{i} = l_{id}+\lambda l_{in},
\end{equation}
where $\lambda$ is a trade-off parameter, which is set to 1 in all experiments.

\subsection{Occlusion-aware RefineNet}
\label{refinenet}
\begin{figure}[t]
    \centering
    \includegraphics[width=\linewidth]{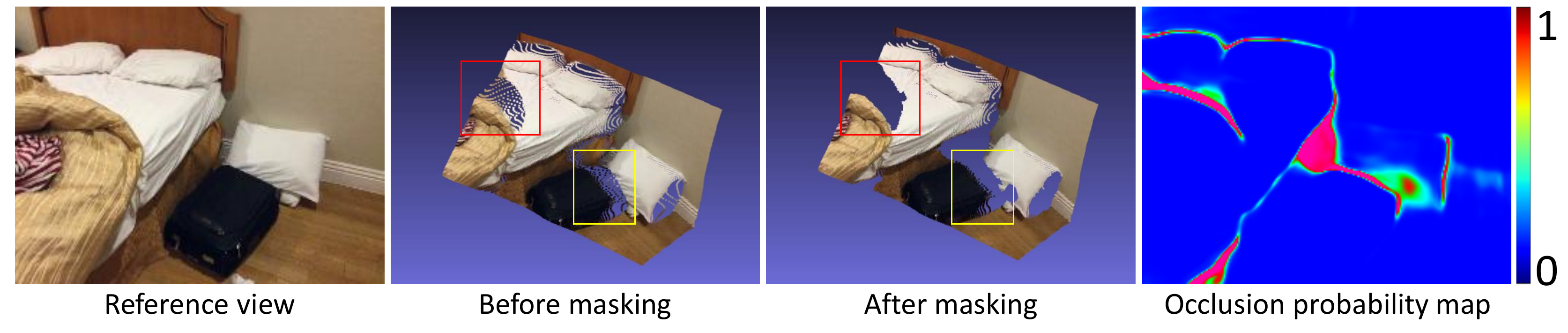}
    \caption{Efficiency of occlusion probability map. Masked by the occlusion map, the point cloud converted from estimated depth has fewer outlying points.}
    \label{fig:occlusion_confidence}
\end{figure}

The next step is to combine the initial depth maps $D_{ref}^1$ and $D_{ref}^2$ of the reference image predicted from different image pairs $\{I_{ref}, I_{s}^{i} \}_{i=1,2}$ into one final depth map, which is denoted as $D_{fin}$.
We design an occlusion-aware network, namely \emph{RefineNet}, based on an occlusion probability map. Note that the occlusion refers to the region in $I_{ref}$ where it cannot be observed in either $I_{s}^{1}$ or $I_{s}^{2}$. 
In contrast to treating all pixels equally, when calculating the loss we assign the lower weights to the occluded region and the higher weights to the non-occluded region, which shifts the focus of the network to the non-occluded regions, since the depth prediction in the non-occluded regions is more reliable. Furthermore, the occlusion probability map predicted by the network can be used to filter out unreliable depth samples, as shown in Fig.~\ref{fig:occlusion_confidence}, which is useful when the depth maps are fused for 3D reconstruction (see more details in Section~\ref{Ablation}).

Here we describe more technical details of this step. We design the \emph{RefineNet} to predict the final depth map and the occlusion probability map from the average cost volume $\Bar{V}$ of the two cost volumes $\{V_i \}_{i=1,2}$ and two initial depth maps $ \{ D_{ref}^{i}\}_{i=1,2}$. The \emph{RefineNet} has one encoder and two decoders. 

The first decoder estimates the occlusion probabilities based on the occlusion information encoded in the average cost volume $\Bar{V}$ and the initial depth maps. Intuitively, for a pixel in the non-occluded region, it has the strongest response (peak) at $n^{th}$ layer with depth $d_n$ of the average cost volume $\Bar{V}$, and $D_{ref}^{1}$ and $D_{ref}^{2}$ at this pixel have similar depth values. However, for a pixel in the occluded region, it has scattered responses at the depth layers of $\Bar{V}$ and has very different values on the initial depth maps at this pixel. The other decoder predicts the refined depth map with the depth constraint and the CNM constraint, as described in Section~\ref{depthnet}

To train the \emph{RefineNet}, we design a novel occlusion-aware loss function as follows, 
\begin{equation}
\label{refine_loss}
{L}_{refine}=\left (  l_{rd} + \beta \cdot l_{rn}  \right ) 
- \alpha \cdot \frac{1}{\left| \mathcal{Q}\right|}\cdot \sum_{q \in \mathcal{Q}} \left ( 1-{P}(q) \right),
\end{equation}
where
\begin{equation}
{l}_{rd}=\frac{1}{\left| \mathcal{Q} \right|} \sum_{q \in \boldsymbol{Q}} \left ( 1-{P}(q) \right) \left\|\hat{D}(q)-{D}_{r}(q)\right\|_{1}
\end{equation}
\begin{equation}
{l}_{rn} =-\frac{1}{\left| \mathcal{Q} \right|} \sum_{q \in \boldsymbol{Q}}
\left ( 1-{P}(q) \right)\hat{N}(q) \cdot {N}_{r}(q).
\end{equation}
Here ${D}_{r}(q)$ denotes the refined estimated depth at pixel $q$, ${N}_{r}(q)$ denotes the surface normal of the 3D point corresponding to $q$, $\hat{D}(q)$ is the ground truth depth of $q$, $\hat{N}(q)$ is the combined normal of $q$, and ${P}(q)$ denotes the occlusion probability value of $q$ (${P}(q)$ is high when $q$ is occluded). The weight $\alpha$ is set to 0.2 and $\beta$ is set to 1 in all experiments.

\section{Datasets and Implementation Details}
\noindent
{\bf Dataset} We train our network with the ScanNet dataset~\cite{dai2017scannet}, and evaluate our method on the 7scenes dataset~\cite{shotton2013scene} and the SUN3D dataset~\cite{xiao2013sun3d}. ScanNet consists of 1600 different indoor scenes, which are divided into 1000 for training and 600 for testing. ScanNet provides RGB images, ground truth depth maps and camera poses. We generate the {\em CNM} as described in Section~\ref{depthnet}.

\noindent
{\bf Implementation details} Our training process consists of three stages. First, we train the \emph{DepthNet} using the loss function defined in Equation~\ref{depthnetloss}. Then, we fix the parameters of the \emph{DepthNet} and train the \emph{RefineNet} with the loss function Equation~\ref{refine_loss}. Next, we finetune the parameters of the \emph{DepthNet} and the \emph{RefineNet} together with the loss terms in Equation~\ref{depthnetloss} and Equation~\ref{refine_loss}. For all the training stages, the ground truth depth map and the \emph{CNM} are used as supervision. We use Adam optimizer ($lr=0.0001$, $\beta_1=0.9$, $\beta_2=0.999$, $weight\_decay=0.00001$) and run for 6 epochs for each stage. We implemented our model in Pytorch~\cite{paszke2017automatic}. Training the network with two GeForce RTX 2080 Ti GPUs, which takes two days for all stages.

\section{Experiments}
To evaluate our method, we compare our method with state-of-the-arts in three aspects: accuracy of depth estimation, geometric consistency, and video-based 3D reconstruction with TSDF fusion.

\begin{figure*}[t]
    \centering
    \includegraphics[width=\linewidth]{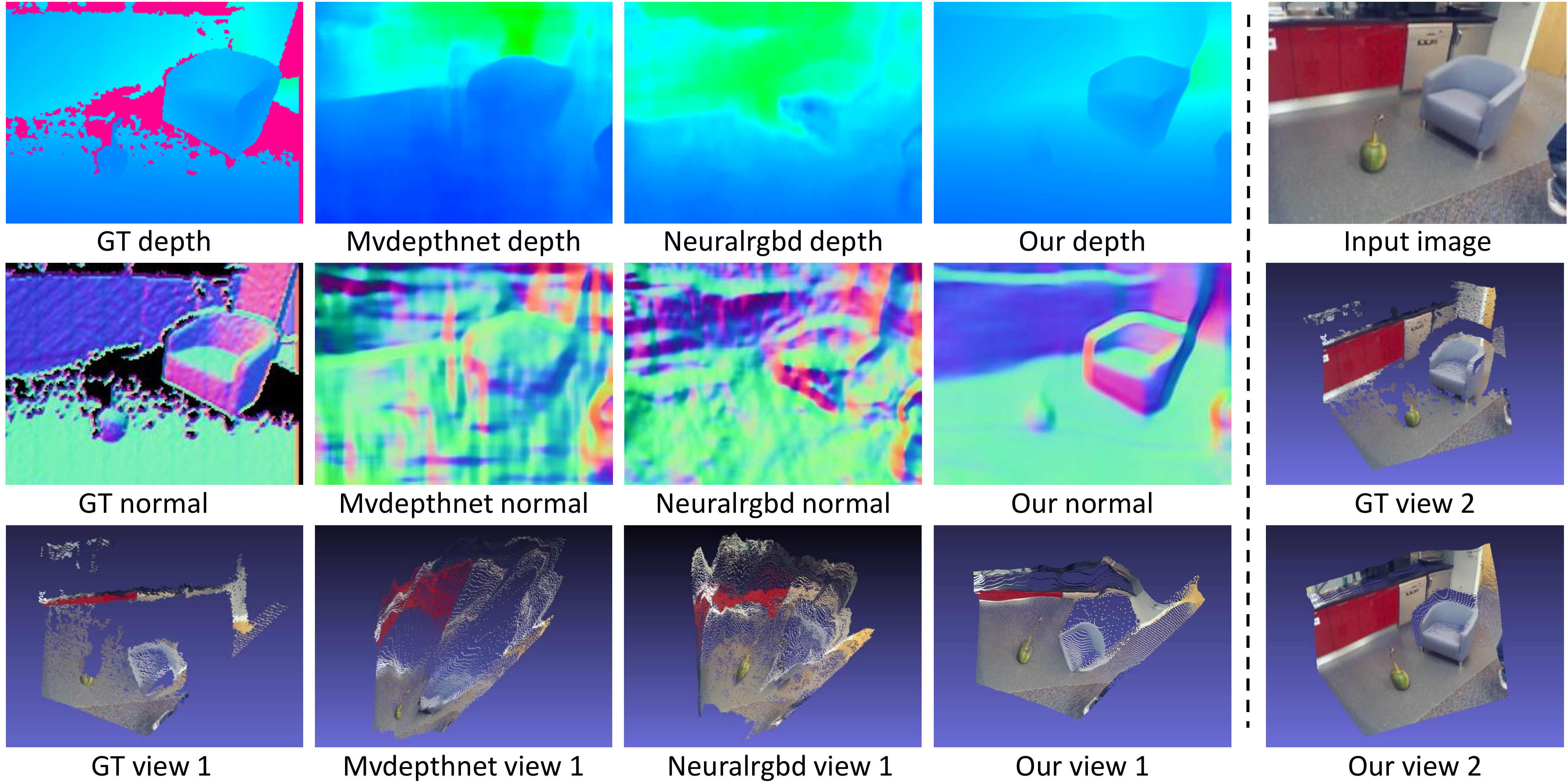}
    \caption{Visual depth comparison with mvdepthnet~\cite{wang2018mvdepthnet} and neuralrgbd~\cite{liu2019neural}. Our estimated depth maps preserve shape regularity better than by mvdepthnet and neuralrgbd, for example, in the regions of sofa, cabinet and wall. The result by mvdepthnet (colum 2) shows strong discontinuity of depth value at the lower end of the red wall, which is incorrect. We also show the visualization of normal maps computed from these depth maps to further show the superior quality of our depth estimation in terms of shape regularity. Note that the red region of GT depth (black region of GT normal) is invalid due to scanning failure of the Kinect sensor. More comparisons can be found in the supplementary materials}
      \label{fig:depth_compare_single}
\end{figure*}

\subsection{Evaluation Metrics}
For depth accuracy, we compute the widely-used statistical metrics defined in~\cite{eigen2014depth}: i) accuracy
under threshold ($\sigma < {1.25}^{i} \text{ where } i\in\{1,2,3\}$); 
ii) scale-invariant error (scale.inv); iii) absolute relative error (abs.rel); iv) square relative error (sq.rel); iv)  root mean square error (rmse); v) rmse in log space (rmse log). 
    
We evaluate the surface normal accuracy with the following metrics used in the prior works~\cite{eigen2015predicting,qi2018geonet}: i) the mean of angle error (mean); ii) the median of the angle error (median); iii) rmse; iv) the percentage of pixels with angle error below threshold $t \text{ where } t \in\left[11.25^{\circ}, 22.5^{\circ}, 30^{\circ}\right]$.

\begin{table}[t]
\begin{center}
\caption{Comparison of depth estimation over 7-Scenes dataset~\cite{shotton2013scene} with the metrics defined in~\cite{eigen2014depth}.}
\resizebox{\linewidth}{!}{%
\begin{threeparttable}[b]
\begin{tabular}{l c c c c c c c c}
\hline
  & $\sigma < 1.25$ & $\sigma < {1.25}^{2}$ & $\sigma < {1.25}^{3}$ & abs.rel & sq.rel & rmse & rmse log & scale.inv \\
\hline
DORN~\cite{fu2018deep} & 60.05 & 87.76 & 96.33 & 0.2000 & 0.1153 & 0.4591 & 0.2813 & 0.2207\\
GeoNet~\cite{qi2018geonet} & 55.10 & 84.46 & 94.76 & 0.2574 & 0.1762 & 0.5253 & 0.3952 & 0.3318\\
Yin \etal~\cite{Yin2019enforcing} & 60.62 & 88.81 & 97.44 & 0.2154 & 0.1245 & 0.4500 & 0.2597 & 0.1660 \\
DeMoN~\cite{ummenhofer2017demon} & 31.88 & 61.02 & 82.52 & 0.3888 & 0.4198 & 0.8549 & 0.4771 & 0.4473\\
{MVSNet-retrain}\tnote{$\dagger$}~\cite{yao2018mvsnet} & 64.09 & 87.73 & 95.88 & 0.2339 & 0.1904 & 0.5078 & 0.2611 & 0.1783\\
{neuralrgbd}\tnote{$\dagger$}~\cite{liu2019neural} & 69.26 & 91.77 & 96.82 & 0.1758 & 0.1123 & 0.4408 & 0.2500 & 0.1899\\
mvdepthnet~\cite{wang2018mvdepthnet} & 71.97  & 92.00 & 97.31 & 0.1865 & 0.1163 & 0.4124 & 0.2256 & 0.1691\\
{mvdepthnet-retrain}\tnote{$\dagger$}~\cite{wang2018mvdepthnet} & 71.79 & 92.56 & 97.83 & 0.1925 & 0.2350 & 0.4585 & 0.2301 & 0.1697\\
DPSNet~\cite{im2019dpsnet} & 63.65 & 85.73 & 94.33 & 0.2710 & 0.2752 & 0.5632 & 0.2759 & 0.1856 \\
{DPSNet-retrain}\tnote{$\dagger$}~\cite{im2019dpsnet}  & 70.96 & 91.42 & 97.17 & 0.1991 & 0.1420 & 0.4382 & 0.2284 & 0.1685 \\
Kusupati \etal\tnote{$\dagger$}~\cite{kusupati2019normal} & 73.12& 92.33 & 97.78&  0.1801 & 0.1179  & 0.4120 & 0.2184 & / \\
{ours}\tnote{$\dagger$} & \bf{76.64} & \bf{94.46} & \bf{98.56} & \bf{0.1612} & \bf{0.0832} & \bf{0.3614} & \bf{0.2049} & \bf{0.1603} \\
\hline
\end{tabular}
\begin{tablenotes}
     \item[$\dagger$] Trained on ScanNet dataset.
   \end{tablenotes}
\end{threeparttable}
}
\label{tab: depth_comparision_state_arts}
\end{center}
\end{table}

\subsection{Comparisons}
\noindent
{\bf Depth prediction} We compare our method with several other depth estimation methods. We categorize them according to their input formats: i) one single image: DORN~\cite{fu2018deep}, GeoNet~\cite{qi2018geonet} and VNL~\cite{Yin2019enforcing}; ii) two images from monocular camera: DeMoN~\cite{ummenhofer2017demon}; iii) multiple unordered images: MVSNet~\cite{yao2018mvsnet}, DPSNet~\cite{im2019dpsnet}, kusupati \etal~\cite{kusupati2019normal}; iv) a video sequence: mvdepthnet~\cite{wang2018mvdepthnet} and neuralrgbd~\cite{liu2019neural}. The models provided by these methods were trained on different datasets and are evaluated on a separate dataset, 7Scenes dataset~\cite{shotton2013scene}. For the multi-view depth estimation methods, neuralrgbd~\cite{liu2019neural} and kusupati \etal~\cite{kusupati2019normal} are trained on ScanNet, MVSNet~\cite{yao2018mvsnet} is trained on DTU dataset~\cite{aanaes2016large}, mvdepthnet~\cite{wang2018mvdepthnet} and DPSNet~\cite{im2019dpsnet} are trained on mixed datasets (SUN3D~\cite{xiao2013sun3d}, TUM RGB-D~\cite{sturm2012benchmark},
MVS datasets~\cite{ummenhofer2017demon}, SceneNN~\cite{hua2016scenenn}  and synthetic dataset Scenes11~\cite{ummenhofer2017demon}). For fair comparison, we further retrained MVSNet~\cite{yao2018mvsnet}, mvdepthnet~\cite{wang2018mvdepthnet} and DPSNet~\cite{im2019dpsnet} on ScanNet. 
Table~\ref{tab: depth_comparision_state_arts} shows that our method outperforms other methods in terms of all evaluation metrics.

\noindent
{\bf Visual comparison} In Fig.~\ref{fig:depth_compare_single}, compared to mvdepthnet~\cite{wang2018mvdepthnet} and neuralrgbd~\cite{liu2019neural}, our estimated depth map has less noise, sharper boundaries and spatially consistent depth values, which can be also seen in the surface normal visualization. Furthermore, the 3D point cloud exported from the estimated depth preserves global planar features and local features in the high-curvature regions. More comparison examples are included in the supplementary materials.

\noindent
{\bf Surface normal accuracy}
\label{surface_normal_compare}
To evaluate the accuracy of normal calculated from estimated depth, we choose two single-view depth estimation methods: GeoNet~\cite{qi2018geonet} and Yin \etal~\cite{Yin2019enforcing}, and one multi-view depth estimation method: Kusupati \etal~\cite{kusupati2019normal}. GeoNet~\cite{qi2018geonet} incorporates surface normal constraint to depth estimation, while Yin \etal~\cite{Yin2019enforcing} propose a global normal constraint, namely {\em Virtual Normal}, which is defined by three randomly sampled non-collinear points. Kusupati \etal~\cite{kusupati2019normal} use the sobel operator to calculate depth gradient and enforce the consistency of the depth gradient and the normal in the pixel coordinate space. To demonstrate that the improved performance of our method indeed benefits from the new CNM constraint (rather than entirely due to our multi-view input), we also retrained our network by replacing the Combined Normal Loss with the Virtual Normal loss (denoted as ``Ours with VNL" in Table~\ref{tab: normal_compare_with_geonet}) on ScanNet dataset.

\begin{figure}[!tp]
    \centering
    \includegraphics[width=\linewidth]{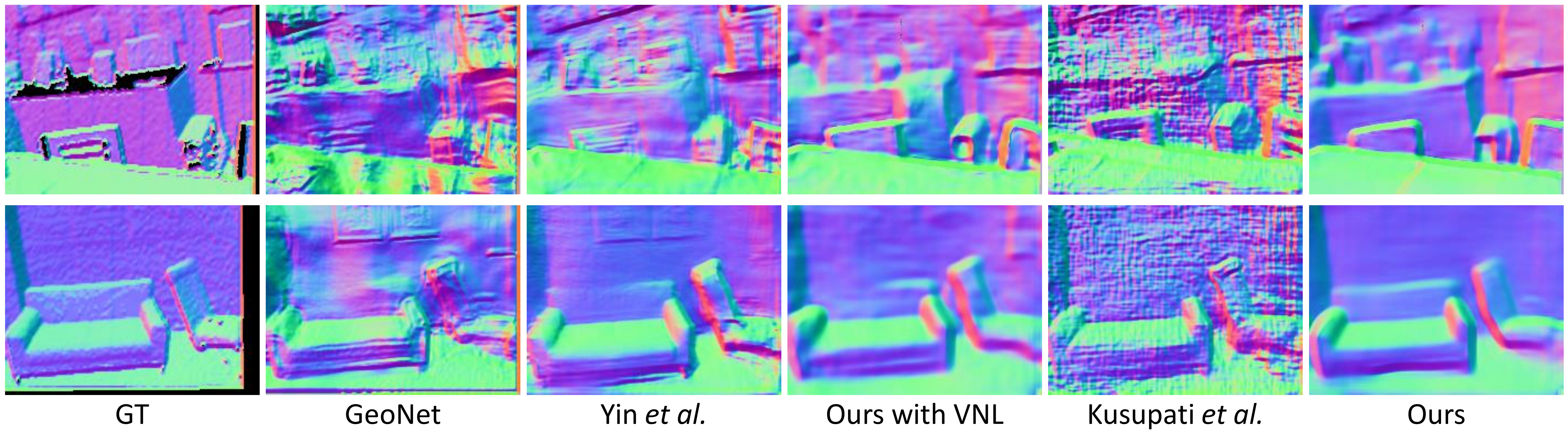}
    \caption{Visual comparison of surface normal calculated from 3D point cloud exported from estimated depth with GeoNet~\cite{qi2018geonet} , Yin \etal~\cite{Yin2019enforcing} and Kusupati \etal~\cite{kusupati2019normal}.}
    \label{fig:normal_compare_geonet}
\end{figure}

As shown in Table~\ref{tab: normal_compare_with_geonet} and Fig.~\ref{fig:normal_compare_geonet}, our method outperforms GeoNet~\cite{qi2018geonet}, Yin \etal~\cite{Yin2019enforcing} and Kusupati \etal~\cite{kusupati2019normal} both quantitatively and qualitatively. Compared with our model retrained using VNL, our model with the {\em CNM} constraint works much better, which preserves local and global features.

\begin{table*}[t]
\begin{center}
\caption{Evaluation of the calculated surface normal from estimated depth on 7scenes\cite{shotton2013scene} and SUN3D~\cite{xiao2013sun3d}.``Ours with VNL" denotes our model retrained using virtual normal loss~\cite{Yin2019enforcing}.}
\resizebox{0.99\linewidth}{!}{%
\begin{tabular}{l ccc|ccc ||ccc|ccc}
\hline
Dataset & \multicolumn{6}{c}{7scenes} & \multicolumn{6}{c}{SUN3D} \\
\hline
  & \multicolumn{3}{c|}{error} & \multicolumn{3}{c||}{accuracy } & \multicolumn{3}{c|}{error } & \multicolumn{3}{c}{accuracy}\\
Method & mean & median & rmse & $ 11.25^{\circ}$ & $22.5^{\circ}$ & $30^{\circ}$ & mean & median & rmse & $ 11.25^{\circ}$ & $22.5^{\circ}$ & $30^{\circ}$\\
\hline
GeoNet~\cite{qi2018geonet} & 57.04 & 52.43 & 65.50 & 4.396 & 15.16 & 23.83 & 47.39 & 40.25 & 57.11 & 9.311 & 26.63 & 37.49\\
Yin \etal~\cite{Yin2019enforcing} & 45.17 & 37.77 & 54.77 & 11.05 & 29.76 & 40.78 & 38.50 & 30.03 & 48.90 & 19.06 & 40.32 & 51.17\\
Ours with VNL & 43.08 & 34.98 & 53.15 & 13.84 & 33.37 & 43.90 & 31.68 & 22.10 & 42.74 & 29.50 & 52.97 & 62.81\\
Kusupati \etal~\cite{kusupati2019normal} & 55.54 & 50.46 & 64.28 & 5.774 & 18.03 & 26.81 & 52.86 & 47.03& 62.30 & 7.507 & 22.56 & 32.22 \\
Ours & {\bf 36.34} & {\bf 27.21} & {\bf 46.46} & {\bf 20.24} & {\bf 43.36} & {\bf 54.11} & {\bf 29.21} & {\bf 20.89} & {\bf 38.99} & {\bf 30.45} & {\bf 55.05} & {\bf 65.25}\\
\hline
\end{tabular}%
}
\label{tab: normal_compare_with_geonet}
\end{center}
\end{table*}

\subsection{Video Reconstruction} 
\begin{figure*}[!tp]
    \centering
    \includegraphics[width=\textwidth]{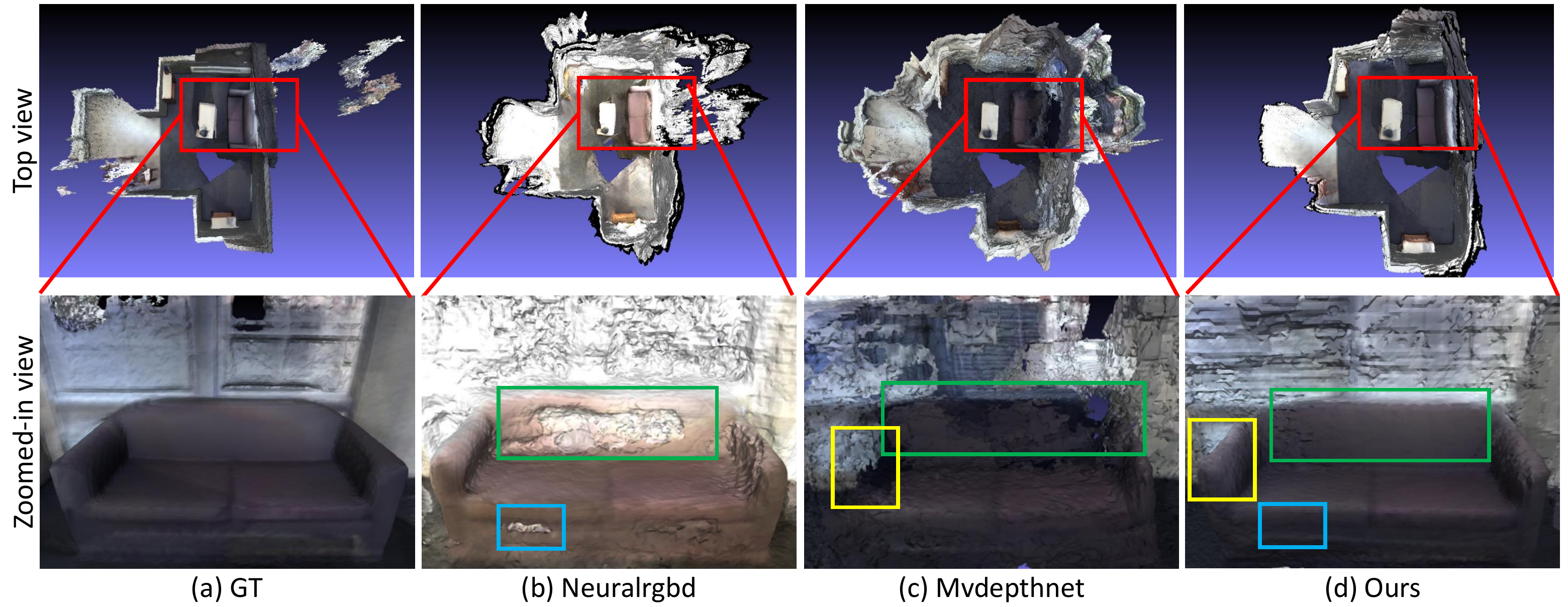}
    \caption{Comparison with neuralrgbd~\cite{liu2019neural} and mvdepthnet~\cite{wang2018mvdepthnet} for 3D reconstruction on a scene from ScanNet. (a) With ground truth depth; (b) With estimated depth and confidence map from neuralrgbd; (c) With estimated depth from mvdepthnet; (d) With our estimated depth and occlusion probability map.  All reconstructions are done with TSDF fusion~\cite{zeng20163dmatch} using 110 frames uniformly selected from a video of 1100 frames.}
    \label{fig:video_recon}
\end{figure*}

With the high-fidelity depth map and the occlusion probability map obtained by our method, high-quality reconstruction of video even in texture-less environments can be achieved by applying TSDF fusion method~\cite{zeng20163dmatch}. We compare our method with mvdepthnet~\cite{wang2018mvdepthnet} and neuralrgbd~\cite{liu2019neural}. Note that these three methods are all trained on the ScanNet dataset. 
As shown in Fig.~\ref{fig:video_recon}, even for white walls, feature-less sofa and table, our reconstructed result is much better than the other two methods in the aspects of local and global structural recovery, and completion. Also, the color of our reconstruction is closest to the color of the ground truth. This confirms that the individual depth maps have less noise and high-consistency so the colors would not be mixed up in the fusion process.

\subsection{Ablation studies}
\label{Ablation}
Compared with prior works, we explicitly drive the depth estimation network to adaptively learn local surface directions and global planar structures. Taking multiple views as inputs, our refinement module jointly refines depth and predicts occlusion probability. In this section, we evaluate the usefulness of each component.

\begin{table*}[t]
\begin{center}
\caption{Evaluation of the usefulness of \emph{CNM} and the occlusion-aware refinement module. The first four rows show the results without any constraint, with only local normal constraint, with only global normal constraint, and with the \emph{CNM} constraint respectively. The last two rows show the results without/with the occlusion-aware loss.}
\resizebox{\linewidth}{!}{%
\begin{tabular}{c c c c|| c c| c c c c ||  c c| c c c}
\hline
\multicolumn{4}{c||}{Components}& \multicolumn{6}{c||}{Estimated depth evaluation} & \multicolumn{5}{c}{Calculated normal evaluation} \\ 
\hline
 local & global & refine & occlu. & $1.25$ & ${1.25}^{2}$  & abs.rel & sq.rel & rmse  & scale.inv & $11.25^{\circ}$ & $22.5^{\circ}$ & mean & median & rmse  \\
\hline
$\times$ & $\times$ & $\times$ & $\times$ & 71.79 & 92.56  & 0.1925 & 0.2350 & 0.4585  & 0.1697 & 9.877 & 27.20 & 46.62 & 39.91 & 55.96 \\
\checkmark & $\times$ & $\times$ & $\times$ & 71.95 & 92.45  & 0.1899 & 0.1188 & 0.4060  & 0.1589 & 16.97 & 39.49 & 37.92 & 29.60 & 47.53 \\
$\times$ & \checkmark &  $\times$ & $\times$ & 66.85 & 90.17 & 0.2096 & 0.1462 & 0.4570 & 0.1752 & 15.39 & 37.09 & 39.15 & 31.33 & 48.45 \\
\checkmark & \checkmark & $\times$ & $\times$ & 73.17 & 92.75  &  0.1812 & 0.1076 & 0.3952  & 0.1654 & 17.67 & 40.23 & 37.65 & 29.18 & 47.36 \\
\checkmark & \checkmark & \checkmark& $\times$ &74.77 & 93.22 & 0.1726 & 0.0999 & 0.3877 & 0.1758 & 18.61 & 41.13 & 37.67 & 28.77 & 47.74 \\
\checkmark & \checkmark & \checkmark & \checkmark & \bf{75.80} & \bf{93.79}  & \bf{0.1669} & \bf{0.0909} & \bf{0.3731}  & \bf{0.1638} & \bf{20.12} & \bf{42.76} & \bf{36.82} & \bf{27.67} & \bf{47.03} \\
\hline
\end{tabular}
}
\label{tab: lgnormal_refine_depth_normal}
\end{center}
\end{table*}

\begin{figure}[!tp]
    \centering
    \includegraphics[width=\linewidth]{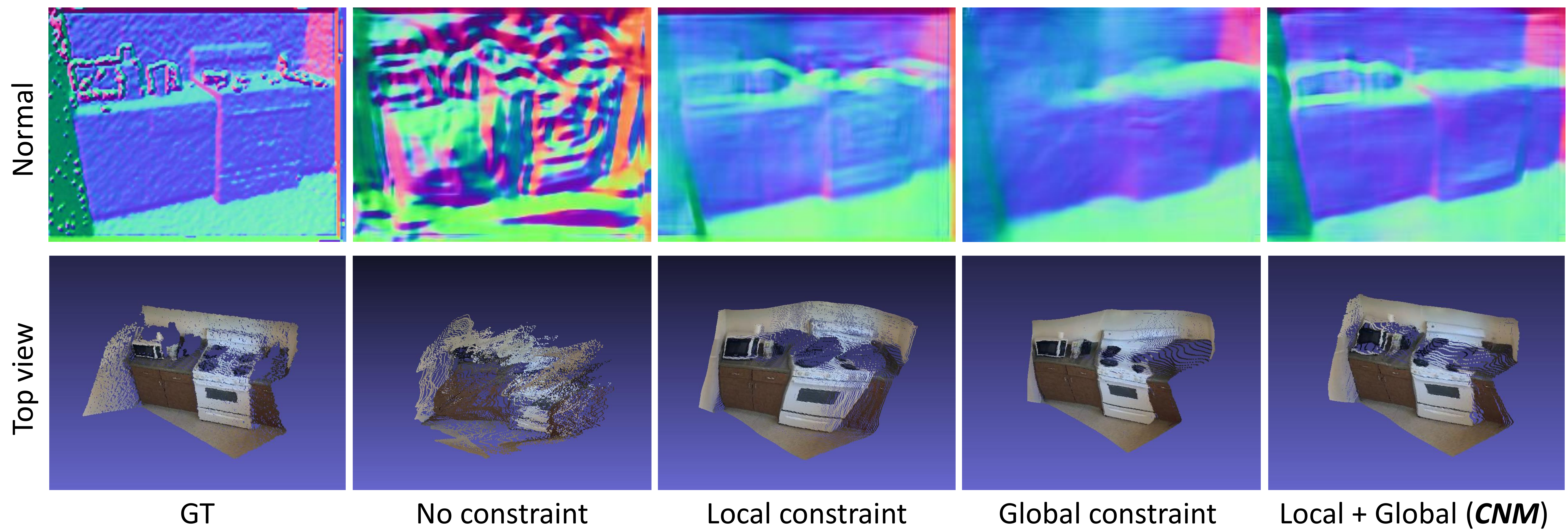}
    \caption{Effects of using local surface normal and \emph{CNM} as constraints for supervision.}
    \label{fig:lgnormal_compare}
\end{figure}

\noindent
{\bf Combined normal map} As shown in Table~\ref{tab: lgnormal_refine_depth_normal}, enforced by our \emph{CNM} constraint (local + global), our model achieves better performance in terms of estimated depth and calculated normal from estimated depth, compared to that without geometric constraint, only with local normal constraint, and only with global normal constraint (the plane normal constraint). Fig.~\ref{fig:lgnormal_compare} also shows that with \emph{CNM}, the calculated normal is more consistent in planar regions and the point cloud converted from estimated depth keeps better shape.

\begin{table}[t]
\begin{center}
\caption{Effect of the number of source views $N$ and the choice of reference view. We use the model trained with $N=2$, and test with different number of views over 7-Scenes dataset~\cite{shotton2013scene}. The first row shows the result of using the last frame in the local window as the reference frame. The other rows are the results of using the middle frame as the reference frame.}
\resizebox{0.95\linewidth}{!}{%
\begin{tabular}{c c c| c c c || c c | c c c}
\hline
  & \multicolumn{5}{c}{Estimated depth evaluation} & \multicolumn{5}{c}{Calculated normal evaluation} \\
\hline
view num & $1.25$ & ${1.25}^{2}$  & abs.rel  & rmse  & scale.inv  & $11.25^{\circ}$ & $22.5^{\circ}$ & mean & median & rmse\\
\hline
2 views (last) & 72.88 & 92.70 & 0.1750 &  0.3910 & 0.1686 & 19.14 & 41.83 & 37.17 & 28.27 & 47.20\\
2 views & 75.80 & 93.79 & 0.1669  & 0.3731 & 0.1638 & 20.12 & 42.76 & 36.82 & 27.67 & 47.03\\
4 views & 76.29 & 94.28 & 0.1647  & 0.3652 & 0.1608 & 20.30 & 43.17 & 36.57 & 27.37 & 46.78\\
6 views & 76.64 & 94.46 & 0.1612 & 0.3614 & 0.1602 & 20.37 & 43.14 & 36.53 & 27.37 & 46.72\\
\hline
\end{tabular}
}
\label{tab: view_num}
\end{center}
\end{table}

\noindent
{\bf The number of source views for refinement} Our refinement module can allow any arbitrary even number of source views as input, and generate a refined depth map and an occlusion probability map. In Table~\ref{tab: lgnormal_refine_depth_normal}, with refinement module, the performance of our model has been significantly improved. As shown in Table~\ref{tab: view_num}, the quality of refined depth will be improved gradually with the increase of source views. A forward pass with 2/4/6 source views all takes nearly 0.02s on a GeForce RTX 2080 Ti GPU.
Furthermore, we evaluate how the performance is affected by the choice of the reference view. 
We find that the performance of our model will degrade significantly if we use the last frame as the reference view rather than the middle frame in the local time window. This is because the middle frame shares more overlapping areas than the last frame with the other frames in the time window.

\begin{figure}[t]
    \centering
    \includegraphics[width=\linewidth]{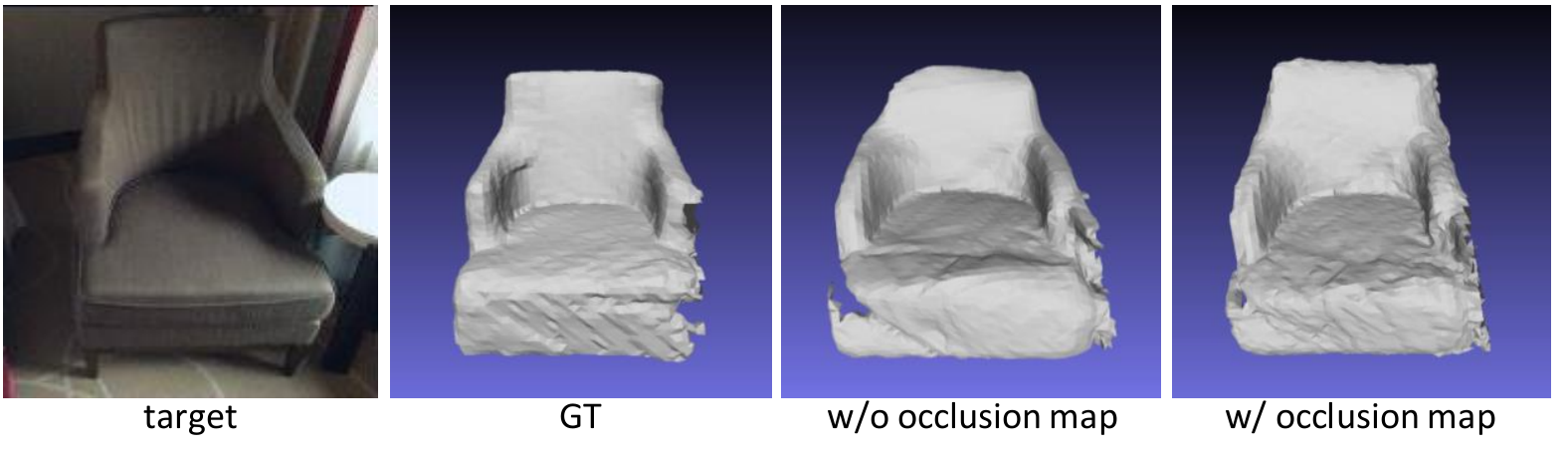}
    \caption{Effect of occlusion probability map for video reconstruction. The occlusion probability map provides weighting parameters into TSDF fusion system for fusing estimated depth maps, and enables reconstructed model to keep sharp boundaries.}
    \label{fig:occlusion_video_recon}
\end{figure}

\noindent
{\bf The usefulness of occlusion probability map} Unlike left-right check as a post-processing operation, our \emph{RefineNet} jointly refines depth and generates an occlusion probability map. As shown in Table.~\ref{tab: lgnormal_refine_depth_normal}, our model with the occlusion-aware refinement achieves better results than that with a naive refinement (w/o occlusion map) and without any refinement. The naive refinement treats every pixel equally regardless whether it is on the occluded region or non-occluded region. In contrast, our occlusion-aware refinement penalizes occluded pixels less and pays more attention to the non-occluded region.
In Fig.~\ref{fig:occlusion_confidence}, the point cloud converted from estimated depth becomes cleaner and has sharp boundaries by applying the occlusion probability map to the multi-view depth integration process. Moreover unlike binary mask, our occlusion map can be easily used in TSDF fusion system~\cite{zeng20163dmatch} as weights. In Fig.~\ref{fig:occlusion_video_recon}, the fused mesh using estimated depth and the occlusion probability map has less artifact than that without occlusion map.

\section{Conclusion and Limitations}
In this paper, we propose a new method for estimating depth from a video sequence. There are two main contributions. We propose a novel normal-based constraint \emph{CNM}, which is designed to preserve local and global geometric features for depth estimation, and a new occlusion-aware strategy to aggregate multiple depth predictions into one single depth map. Experiments demonstrate that our method outperforms than the state-of-the-art in terms of the accuracy of depth prediction and the recovery of geometric features. 
Furthermore, the high-fidelity depth prediction and the occlusion detection make the highly-detailed reconstruction with only a commercial RGB camera possible.

Now we discuss the limitations of our work and possible future directions. First, the performance of our method relies on the quality of the \emph{CNM}, which is based on the segmentation of global planar regions. It has been observed that existing plane segmentation methods are not robust for all the scenes. One possible solution is to jointly learn the segmentation labels and the depth prediction. Second, for the task of video-based 3D reconstruction, our work can be extended by designing an end-to-end network to directly generate 3D reconstruction from video, rather than having to invoke the explicit step of using TSDF fusion to integrate the estimated depth maps. 

\begin{acks}
We thank the anonymous reviewers for their valuable feedback. Wenping Wang acknowledges support from AIR@InnoHK -- Center for Transformative Garment Production (TransGP). Christian Theobalt acknowledges support from ERC Consolidator Grant 4DReply (770784). Lingjie Liu acknowledges support from Lise Meitner Postdoctoral Fellowship. 
\end{acks}

%
%
\bibliographystyle{splncs04}

\begin{thebibliography}{10}
\providecommand{\url}[1]{\texttt{#1}}
\providecommand{\urlprefix}{URL }
\providecommand{\doi}[1]{https://doi.org/#1}

\bibitem{humayun2011learning}
Learning to find occlusion regions. In: CVPR 2011. pp. 2161--2168. IEEE (2011)

\bibitem{aanaes2016large}
Aan{\ae}s, H., Jensen, R.R., Vogiatzis, G., Tola, E., Dahl, A.B.: Large-scale
  data for multiple-view stereopsis. International Journal of Computer Vision
  \textbf{120}(2),  153--168 (2016)

\bibitem{alvarez2016collision}
Alvarez, H., Paz, L.M., Sturm, J., Cremers, D.: Collision avoidance for
  quadrotors with a monocular camera. In: Experimental Robotics. pp. 195--209.
  Springer (2016)

\bibitem{bleyer2011patchmatch}
Bleyer, M., Rhemann, C., Rother, C.: Patchmatch stereo-stereo matching with
  slanted support windows. In: Bmvc. vol.~11, pp. 1--11 (2011)

\bibitem{chang2018pyramid}
Chang, J.R., Chen, Y.S.: Pyramid stereo matching network. In: Proceedings of
  the IEEE Conference on Computer Vision and Pattern Recognition. pp.
  5410--5418 (2018)

\bibitem{dai2017scannet}
Dai, A., Chang, A.X., Savva, M., Halber, M., Funkhouser, T., Nie{\ss}ner, M.:
  Scannet: Richly-annotated 3d reconstructions of indoor scenes. In:
  Proceedings of the IEEE Conference on Computer Vision and Pattern
  Recognition. pp. 5828--5839 (2017)

\bibitem{egnal2002detecting}
Egnal, G., Wildes, R.P.: Detecting binocular half-occlusions: Empirical
  comparisons of five approaches. IEEE Transactions on pattern analysis and
  machine intelligence  \textbf{24}(8),  1127--1133 (2002)

\bibitem{eigen2015predicting}
Eigen, D., Fergus, R.: Predicting depth, surface normals and semantic labels
  with a common multi-scale convolutional architecture. In: Proceedings of the
  IEEE international conference on computer vision. pp. 2650--2658 (2015)

\bibitem{eigen2014depth}
Eigen, D., Puhrsch, C., Fergus, R.: Depth map prediction from a single image
  using a multi-scale deep network. In: Advances in neural information
  processing systems. pp. 2366--2374 (2014)

\bibitem{fu2018deep}
Fu, H., Gong, M., Wang, C., Batmanghelich, K., Tao, D.: Deep ordinal regression
  network for monocular depth estimation. In: Proceedings of the IEEE
  Conference on Computer Vision and Pattern Recognition. pp. 2002--2011 (2018)

\bibitem{Furukawa:2015:MST:2864699.2864700}
Furukawa, Y., Hern\'{a}ndez, C.: Multi-view stereo: A tutorial. Found. Trends.
  Comput. Graph. Vis.  \textbf{9}(1-2),  1--148 (Jun 2015).
  \doi{10.1561/0600000052}, \url{http://dx.doi.org/10.1561/0600000052}

\bibitem{gallup2007real}
Gallup, D., Frahm, J.M., Mordohai, P., Yang, Q., Pollefeys, M.: Real-time
  plane-sweeping stereo with multiple sweeping directions. In: 2007 IEEE
  Conference on Computer Vision and Pattern Recognition. pp.~1--8. IEEE (2007)

\bibitem{hosni2009local}
Hosni, A., Bleyer, M., Gelautz, M., Rhemann, C.: Local stereo matching using
  geodesic support weights. In: 2009 16th IEEE International Conference on
  Image Processing (ICIP). pp. 2093--2096. IEEE (2009)

\bibitem{hosni2012fast}
Hosni, A., Rhemann, C., Bleyer, M., Rother, C., Gelautz, M.: Fast cost-volume
  filtering for visual correspondence and beyond. IEEE Transactions on Pattern
  Analysis and Machine Intelligence  \textbf{35}(2),  504--511 (2012)

\bibitem{hua2016scenenn}
Hua, B.S., Pham, Q.H., Nguyen, D.T., Tran, M.K., Yu, L.F., Yeung, S.K.:
  Scenenn: A scene meshes dataset with annotations. In: 2016 Fourth
  International Conference on 3D Vision (3DV). pp. 92--101. IEEE (2016)

\bibitem{huang2018deepmvs}
Huang, P.H., Matzen, K., Kopf, J., Ahuja, N., Huang, J.B.: Deepmvs: Learning
  multi-view stereopsis. In: Proceedings of the IEEE Conference on Computer
  Vision and Pattern Recognition. pp. 2821--2830 (2018)

\bibitem{ilg2018occlusions}
Ilg, E., Saikia, T., Keuper, M., Brox, T.: Occlusions, motion and depth
  boundaries with a generic network for disparity, optical flow or scene flow
  estimation. In: Proceedings of the European Conference on Computer Vision
  (ECCV). pp. 614--630 (2018)

\bibitem{im2019dpsnet}
Im, S., Jeon, H.G., Lin, S., Kweon, I.S.: Dpsnet: end-to-end deep plane sweep
  stereo. arXiv preprint arXiv:1905.00538  (2019)

\bibitem{kang2001handling}
Kang, S.B., Szeliski, R., Chai, J.: Handling occlusions in dense multi-view
  stereo. In: Proceedings of the 2001 IEEE Computer Society Conference on
  Computer Vision and Pattern Recognition. CVPR 2001. vol.~1, pp.~I--I. IEEE
  (2001)

\bibitem{Kendall_2017_ICCV}
Kendall, A., Martirosyan, H., Dasgupta, S., Henry, P., Kennedy, R., Bachrach,
  A., Bry, A.: End-to-end learning of geometry and context for deep stereo
  regression. In: The IEEE International Conference on Computer Vision (ICCV)
  (Oct 2017)

\bibitem{kusupati2019normal}
Kusupati, U., Cheng, S., Chen, R., Su, H.: Normal assisted stereo depth
  estimation. arXiv preprint arXiv:1911.10444  (2019)

\bibitem{li2018megadepth}
Li, Z., Snavely, N.: Megadepth: Learning single-view depth prediction from
  internet photos. In: Proceedings of the IEEE Conference on Computer Vision
  and Pattern Recognition. pp. 2041--2050 (2018)

\bibitem{liu2019neural}
Liu, C., Gu, J., Kim, K., Narasimhan, S.G., Kautz, J.: Neural rgb (r) d
  sensing: Depth and uncertainty from a video camera. In: Proceedings of the
  IEEE Conference on Computer Vision and Pattern Recognition. pp. 10986--10995
  (2019)

\bibitem{Liu_2019_CVPR}
Liu, C., Kim, K., Gu, J., Furukawa, Y., Kautz, J.: Planercnn: 3d plane
  detection and reconstruction from a single image. In: The IEEE Conference on
  Computer Vision and Pattern Recognition (CVPR) (June 2019)

\bibitem{liu2015deep}
Liu, F., Shen, C., Lin, G.: Deep convolutional neural fields for depth
  estimation from a single image. In: Proceedings of the IEEE Conference on
  Computer Vision and Pattern Recognition. pp. 5162--5170 (2015)

\bibitem{luo2016efficient}
Luo, W., Schwing, A.G., Urtasun, R.: Efficient deep learning for stereo
  matching. In: Proceedings of the IEEE Conference on Computer Vision and
  Pattern Recognition. pp. 5695--5703 (2016)

\bibitem{min2008cost}
Min, D., Sohn, K.: Cost aggregation and occlusion handling with wls in stereo
  matching. IEEE Transactions on Image Processing  \textbf{17}(8),  1431--1442
  (2008)

\bibitem{paszke2017automatic}
Paszke, A., Gross, S., Chintala, S., Chanan, G., Yang, E., DeVito, Z., Lin, Z.,
  Desmaison, A., Antiga, L., Lerer, A.: Automatic differentiation in pytorch
  (2017)

\bibitem{qi2018geonet}
Qi, X., Liao, R., Liu, Z., Urtasun, R., Jia, J.: Geonet: Geometric neural
  network for joint depth and surface normal estimation. In: Proceedings of the
  IEEE Conference on Computer Vision and Pattern Recognition. pp. 283--291
  (2018)

\bibitem{qiu2019deeplidar}
Qiu, J., Cui, Z., Zhang, Y., Zhang, X., Liu, S., Zeng, B., Pollefeys, M.:
  Deeplidar: Deep surface normal guided depth prediction for outdoor scene from
  sparse lidar data and single color image. In: Proceedings of the IEEE
  Conference on Computer Vision and Pattern Recognition. pp. 3313--3322 (2019)

\bibitem{Seitz:2006:CEM:1153170.1153518}
Seitz, S.M., Curless, B., Diebel, J., Scharstein, D., Szeliski, R.: A
  comparison and evaluation of multi-view stereo reconstruction algorithms. In:
  Proceedings of the 2006 IEEE Computer Society Conference on Computer Vision
  and Pattern Recognition - Volume 1. pp. 519--528. CVPR '06, IEEE Computer
  Society, Washington, DC, USA (2006). \doi{10.1109/CVPR.2006.19},
  \url{http://dx.doi.org/10.1109/CVPR.2006.19}

\bibitem{shotton2013scene}
Shotton, J., Glocker, B., Zach, C., Izadi, S., Criminisi, A., Fitzgibbon, A.:
  Scene coordinate regression forests for camera relocalization in rgb-d
  images. In: Proceedings of the IEEE Conference on Computer Vision and Pattern
  Recognition. pp. 2930--2937 (2013)

\bibitem{sturm2012benchmark}
Sturm, J., Engelhard, N., Endres, F., Burgard, W., Cremers, D.: A benchmark for
  the evaluation of rgb-d slam systems. In: 2012 IEEE/RSJ International
  Conference on Intelligent Robots and Systems. pp. 573--580. IEEE (2012)

\bibitem{ummenhofer2017demon}
Ummenhofer, B., Zhou, H., Uhrig, J., Mayer, N., Ilg, E., Dosovitskiy, A., Brox,
  T.: Demon: Depth and motion network for learning monocular stereo. In:
  Proceedings of the IEEE Conference on Computer Vision and Pattern
  Recognition. pp. 5038--5047 (2017)

\bibitem{wang2019local}
Wang, J., Zickler, T.: Local detection of stereo occlusion boundaries. In:
  Proceedings of the IEEE Conference on Computer Vision and Pattern
  Recognition. pp. 3818--3827 (2019)

\bibitem{wang2018mvdepthnet}
Wang, K., Shen, S.: Mvdepthnet: real-time multiview depth estimation neural
  network. In: 2018 International Conference on 3D Vision (3DV). pp. 248--257.
  IEEE (2018)

\bibitem{xiao2005motion}
Xiao, J., Shah, M.: Motion layer extraction in the presence of occlusion using
  graph cuts. IEEE transactions on pattern analysis and machine intelligence
  \textbf{27}(10),  1644--1659 (2005)

\bibitem{xiao2013sun3d}
Xiao, J., Owens, A., Torralba, A.: Sun3d: A database of big spaces
  reconstructed using sfm and object labels. In: Proceedings of the IEEE
  International Conference on Computer Vision. pp. 1625--1632 (2013)

\bibitem{xu2008stereo}
Xu, L., Jia, J.: Stereo matching: An outlier confidence approach. In: European
  Conference on Computer Vision. pp. 775--787. Springer (2008)

\bibitem{yang2012non}
Yang, Q.: A non-local cost aggregation method for stereo matching. In: 2012
  IEEE Conference on Computer Vision and Pattern Recognition. pp. 1402--1409.
  IEEE (2012)

\bibitem{yao2018mvsnet}
Yao, Y., Luo, Z., Li, S., Fang, T., Quan, L.: Mvsnet: Depth inference for
  unstructured multi-view stereo. In: Proceedings of the European Conference on
  Computer Vision (ECCV). pp. 767--783 (2018)

\bibitem{Yin2019enforcing}
Yin, W., Liu, Y., Shen, C., Yan, Y.: Enforcing geometric constraints of virtual
  normal for depth prediction. In: The IEEE International Conference on
  Computer Vision (ICCV) (2019)

\bibitem{yoon2006adaptive}
Yoon, K.J., Kweon, I.S.: Adaptive support-weight approach for correspondence
  search. IEEE transactions on pattern analysis \& machine intelligence (4),
  650--656 (2006)

\bibitem{zbontar2015computing}
Zbontar, J., LeCun, Y.: Computing the stereo matching cost with a convolutional
  neural network. In: Proceedings of the IEEE conference on computer vision and
  pattern recognition. pp. 1592--1599 (2015)

\bibitem{zeng20163dmatch}
Zeng, A., Song, S., Nie{\ss}ner, M., Fisher, M., Xiao, J., Funkhouser, T.:
  3dmatch: Learning local geometric descriptors from rgb-d reconstructions. In:
  CVPR (2017)

\bibitem{zhang2018deepdepth}
Zhang, Y., Funkhouser, T.: Deep depth completion of a single rgb-d image. The
  IEEE Conference on Computer Vision and Pattern Recognition (CVPR)  (2018)

\bibitem{Zollhoefer2018RecoSTAR}
{Zollh{\"o}fer}, M., {Stotko}, P., {G{\"o}rlitz}, A., {Theobalt}, C.,
  Nie{\ss}ner, M., {Klein}, R., {Kolb}, A.: {State of the Art on 3D
  Reconstruction with RGB-D Cameras}. Computer Graphics Forum (Eurographics
  State of the Art Reports 2018)  \textbf{37}(2) (2018)

\end{thebibliography}

\end{document}